\documentclass{llncs}

\usepackage{graphicx}
\usepackage{amssymb}
\usepackage{amsmath}
\usepackage{url}
\usepackage{pgfgantt}
\usepackage{todonotes}

\usepackage[shortcuts,acronym]{glossaries}
\usepackage{geometry} 
\usepackage{xcolor}
\usepackage{makecell}

\usepackage{cite}

\usepackage{lipsum}

\usepackage{ifthen}
\newboolean{article} 

\makeglossaries % generates the acronym list

\newacronym{ML}{ML}{machine learning}
\newacronym{OC}{OC}{organic computing}
\newacronym{TL}{TL}{transfer learning}
\newacronym{MTL}{MTL}{multi-task-learning}
\newacronym{IL}{IL}{inductive learning}
\newacronym{TDL}{TDL}{transductive learning}
\newacronym{NLP}{NLP}{natural language processing}
\newacronym{NWP}{NWP}{numerical weather prediction}
\newacronym{CSGE}{CSGE}{coopetitive soft-gating ensemble}
\newacronym{MSDA}{MSDA}{multi source domain adaption}

\newcommand\SEC[1]{Section~\ref{#1}} % replace with Sec. for a short version
\newcommand\FIG[1]{Figure~\ref{#1}} % replace with Fig. for a short version
\newcommand\TBL[1]{Table~\ref{#1}} % replace with TBbl. for a short version

\newcommand\wpt{Challenge} % def for title
\newcommand\wpu{Challenge} % for upper case at the beignning of the sentence
\newcommand\wpl{challenge} % lower case
\newcommand\atl{article~}

\usepackage{fancyhdr}
\begin{document}

\setboolean{article}{true}

\title{Transfer Learning in the Field of Renewable Energies \\ \small A Transfer Learning Framework Providing Power Forecasts Throughout the Lifecycle of Wind Farms After Initial Connection to the Electrical Grid}

\author{Jens Schreiber}

\institute{Intelligent Embedded Systems Lab, University of Kassel, Germany\\
\email{jens.schreiber@uni-kassel.de}}

\maketitle

\begin{abstract}
    In recent years, transfer learning gained particular interest in the field of vision and natural language processing. In the research field of vision, e.g., deep neural networks and transfer learning techniques achieve almost perfect classification scores within minutes. Nonetheless, these techniques are not yet widely applied in other domains. Therefore, this \atl~ identifies critical challenges and shows potential solutions for power forecasts in the field of renewable energies. It proposes a framework utilizing transfer learning techniques in wind power forecasts with limited or no historical data. On the one hand, this allows evaluating the applicability of transfer learning in the field of renewable energy. On the other hand, by developing automatic procedures, we assure that the proposed methods provide a framework that applies to domains in organic computing as well.
\end{abstract}

\section{Introduction}

In recent years, \ac{TL} \cite{Weiss2016} gained particular interest in the field of vision and \ac{NLP} \cite{Sogaard2013}. In the research field of vision, e.g., so-called deep neural networks \cite{Goodfellow} are trained on large amounts of data to obtain a computer model that allows classification of almost arbitrary categories in pictures. By training deep neural networks on large datasets, such as ImageNet\footnote{\url{http://www.image-net.org/}, last accessed 02.10.2018}, the model learns a high-dimensional representation of the data. Transferring generic representation to other more specific domains, e.g., the classification of cats and dogs is called domain adaption or transfer learning. More specifically, the field of domain adaption comprises transferring a model or data from a source domain to a target domain, where domain refers to the input data of the model. Typically, we distinguish between source task, the classification in the ImageNet dataset based on the input data; and the target task, the classification of cats and dogs in pictures. \ac{TL} is especially interesting when there is little or no data in the target domain and thereby helps in improving performance \cite{Pan2010, Torrey2009} by making use of the knowledge from the source task and domain. Moreover, this approach allows reducing computational cost by adopting an already well-performing model on a new problem. Despite all these benefits, \ac{TL} methods applied to vision and \ac{NLP} tasks are not yet often used in the area of renewable power forecasts, even though they have various applications.

Traditionally, a physical model was used, in renewable power forecasts, to predict so-called day-ahead estimates of the expected power generation to integrate these volatile energy resources into the electrical grid. The physical wind power model uses rotor area as well as wind speed and air density from \ac{NWP} as inputs to forecast the expected power. However, in various articles, it has been shown that \ac{ML} techniques, such as neural networks, have superior forecast quality compared to the physical model\cite{Gensler2018}. This improvement is, e.g., due to the capability of neural networks to learn the characteristics of a wind farm's location based on the non-linear relationship of the \ac{NWP} input and the historically measured wind power generation. Respectively, the improved forecast quality allows decision makers to devise more confident decisions compared to the physical model.

However, traditional \ac{ML} techniques, such as linear regression, support vector regression, or random forests, are not applicable if there exist limited or no historical wind power data. E.g., consider the situation where a new wind farm is constructed or a night cutoff of the turbines is implemented: In these situations limited or no historical data of the generated power is available to train an \ac{ML} model. So instead we need to fall back to the physical model that requires no training data. This fall back is problematic because decision makers require the excellent forecast performance from \ac{ML} models to devise a confident justification for the electrical grid at all times. Fortunately, \ac{TL} techniques allow using \ac{ML} methods even with little to no data by inferring knowledge from wind farms where considerable knowledge exists.

Correspondingly, the thesis aims to develop a framework that allows utilizing \ac{TL} in wind power forecasts with limited or no historical data in the lifecycle of a wind farm. Therefore, the thesis intends to evaluate the applicability of \ac{TL} in the field of renewable energy by answering the following questions:

\begin{enumerate}
 \item \textbf{How} can historical source domain data and forecast models transfer to the target domain?
  \item \textbf{What} historical source domain data and forecast model is the most suitable to transfer?
  \item \textbf{How} to integrate an increasing amount of historical data into the model for the new wind farm?
\end{enumerate}

\ifthenelse{\boolean{article}}{
By developing automatic procedures that configure the most suitable data selection themselves, the proposed methods provide a framework that will apply to domains in \ac{OC} as well~\cite{Schloer2017}. Furthermore, self-improvement techniques based on the configured and selected data allow a continuous self-optimization framework based on the pool of information. Moreover, the proposed methods are probably applicable to other domains as well. In traffic scenarios, \ac{TL} can be used to continuously improve the forecasts of pedestrian movements based on various sensor data such as cameras, gyroscope, and GPS. Moreover, models developed for pedestrians are, e.g., transferred automatically for bicyclist or classification tasks.
}{}

The remainder of this \atl~ is structured as follows. First, we summarize related work in \SEC{relwork}. Then we detail the research proposal in \SEC{resproposal}. Based on the research proposal we describe the current progress in \SEC{sec:prog_desc}.

\section{Related Work}
\label{relwork}

In this section related work relevant to the research proposal in \SEC{resproposal} is detailed. First, \SEC{definitionTL} gives a general overview of the current research in \ac{TL}. Afterward, we make the connection to current \ac{TL} techniques applied in renewable energies in \SEC{transLearnRen} and introduce papers relevant to the research proposal. 

\subsection{Transfer Learning}
\label{definitionTL}
Transfer learning is the area of research concerning methods allowing the adaption of knowledge from one domain to another. In particular, it is about methods that allow a domain adaption from a source domain $D_S$ to a target domain $D_T$.

A domain is defined by $D = \{\mathcal{X}, P(X)\}$, where $\mathcal{X}$ is the feature space and $P(X)$ is the marginal distribution with $X \in \mathcal{X}$. The task of a domain relates the \ac{NWP} input to the generated power and is defined with $T = \{ \mathcal{Y},~f(\cdot)\}$, where the function $f(\cdot)$ relates the input features $X$ to the generated power $Y,~\text{with}~Y\in~\mathcal{Y}$. Equally to the domain distinction, we differentiate between the source task $T_S$ and the target task $T_T$. Other formulas are also differentiated by the subscript $S$ for the source and $T$ for the target. The above definitions allow describing various domain adaption problems, e.g., inductive learning, which tackles the difference between $D_S$ and $D_T$. A comprehensive overview of definitions with examples is given in \TBL{tab:defs_tl} and \TBL{tab:merge_tl_def}. Based on these tables, the following list provides some of the most important concepts towards solutions of the research questions:

\begin{itemize}
  \item \textbf{Parameter transfer} finds shared parameters between the source and the target domain. These shared parameters are used to transfer knowledge from $D_S$ to $D_T$~\cite{Torrey2009, Pan2010}. One potential method is an \ac{MTL} approach. In \ac{MTL} multiple tasks are learned at once. Consider an \ac{ML} model, which learns the prediction of $3$ offshore wind farms at once. This learning process will allow finding shared parameters that apply to other similar offshore farms. One common \ac{MTL} approach is hard parameter sharing. In hard parameter sharing all hidden layers of a neural network, are shared across all tasks. Each task has several output layers\cite{Ruder2017a, Ruder2017}. Another method is to use finetuning of a neural network. First, in finetuning, a deep neural network is trained on a large set of wind farms. Second, the model's weights are slightly modified (finetuned) to be applicable in the target domain.

  \item \textbf{Feature representation transfer} intends to find a representation of the \ac{NWP} features minimizing the differences between the source and the target domains\cite{Pan2010}. An autoencoder\cite{Hu2016} can achieve this generic representation. An autoencoder is an unsupervised learning strategy that first compresses \ac{NWP} features (encode) and then restores the original features from the compressed representation (decode)\cite{Goodfellow}. This encoding and decoding process allows the autoencoder to learn a common feature representation between the source and the target domain. Other approaches also use an \ac{MTL} approach, where a so-called embedding layer is applied to determine a generic representation of the source domain. The learned generic source representation is used to finetune for the additional target task\cite{Long2015}. 

  \item \textbf{Instance transfer} makes use of historical $D_S$, $\mathcal{Y}_S$ and some historical target task data $\mathcal{Y}_T$. By weighting the little historical data $\mathcal{Y}_S$, instance transfer allows making the best use of both datasets concerning the forecast quality in the target domain. One method is the self-training approach \cite{Sogaard2013}. First, the self-training approach artificially creates target task data, by a model trained on the source dataset. Second, training a new model on the target domain and the artificially created target response. By only using artificial target data with high confidence for the prediction, it is guaranteed that the data is re-weighted in a way, that just data points contribute to the model that are relevant in the target domain or one could say are transferable.
\end{itemize}

\vspace{-1em}
\begin{table}[h!]
  \begin{center}
    \caption{Terminologies definitions of transfer learning. "-" indicates that the definition does not make any assumption about the domain or the task.}
    \label{tab:defs_tl}
    \begin{tabular}{l|c|c} % <-- Alignments: 1st column left, 2nd middle and 3rd right, with vertical lines in between
      \textbf{} & \textbf{Domain} & \textbf{Task}\\
      %$\alpha$ & $\beta$ & $\gamma$ \\
      \hline
      Inductive Learning \cite{Torrey2009, Pan2010}& - &  $T_S \neq T_T$ \\
      Transduchtive Learning \cite{Torrey2009, Pan2010}& $D_S \neq D_T$  &  $T_S = T_T$ \\
      Homogenous Learning \cite{Weiss2016}& $\mathcal{X_S} = \mathcal{X_T}$ & -\\
      Heterogeneous Learning \cite{Weiss2016}& $\mathcal{X_S} \neq \mathcal{X_T}$ & -\\
    \end{tabular}
  \end{center}
  \vspace{-2.5em}
\end{table}

\vspace{-1em}
\begin{table}[h!]
  \begin{center}
    \caption{To refine the terminology and combine different definitions from \cite{Torrey2009, Pan2010}, and \cite{Weiss2016} the following table gives detailed information about differences in definitions and provides examples in the context of renewable energies.}
    \label{tab:merge_tl_def}
    \begin{tabular}{l||c|l|l} % <-- Alignments: 1st column left, 2nd middle and 3rd right, with vertical lines in between
      \textbf{} & \textbf{Domain} & \textbf{Task}& \textbf{Example}\\
      
      \hline\hline
      \makecell{Homogenous\\Inductive Learning} & $\mathcal{X_S} = \mathcal{X_T}$ & $T_S \neq T_T$ & \makecell[l]{Same weather model and input features are used\\ in the source and target domain, e.g.,\\ wind speed, humidity and wind direction.\\ The source task is adapted to classify iced wind\\ turbines in the target task, instead of forecasting\\ the power generation.}\\
      \hline

      \makecell{Heterogeneous\\Inductive Learning} & $\mathcal{X_S} \neq \mathcal{X_T}$ & $T_S \neq T_T$&\makecell[l]{Source and target domain have a different\\number of weather features, e.g., temperature\\is additionally used in the target domain.\\The source task is adapted to classify iced\\wind turbines in the target task,\\instead of forecasting the power\\generation.}\\
      \hline

      \makecell{Homogenous\\Transductive Learning} & $\mathcal{X_S} = \mathcal{X_T}$, $P(X_S) \neq P(X_T)$ & $T_S = T_T$ & \makecell[l]{Same weather model and input features are used\\ in the source and target domain, e.g.,\\ wind speed, humidity and wind direction.\\ However, two wind farm locations with different\\ marginal distributions of the weather. Source and\\ target task both forecast power generation.}\\

      \hline
      \makecell{Heterogeneous\\ Transductive Learning} & $\mathcal{X_S} \neq \mathcal{X_T}$, $P(X_S) = P(X_T)$ & $T_S = T_T$ & \makecell[l]{Two \ac{NWP} models with related but different\\ features of one wind farm location of the source\\ and the target domain. The underlying marginal\\ distributions of $\mathcal{X_S}$ and $\mathcal{X_T}$ are the same, due to\\ the similar location of the wind farm. Source\\ and target task both forecast power generation.}\\
    \end{tabular}
  \end{center}
  \vspace{-2.5em}
\end{table}

\subsection{Wind Power Forecasting}
\label{transLearnRen}

In this section, we summarize current research on transfer learning in renewable energies. Further on an ensemble technique is introduced that is applicable in the context of transfer learning. 

In~\cite{Hu2016} the authors use transfer learning for wind speed predictions. The authors train a so-called \ac{MTL} deep denoising autoencoder on four wind farms simultaneously. This model allows them to obtain the best forecast results for short time forecast horizons (10-min, 30-min, and 1-h) with a limited amount of data.

The author's of~\cite{Qureshi2017} use a combination of nine autoencoders, deep neural networks, and transfer learning to limit the time required for training. Interestingly, initially, all nine autoencoders are trained based on data from a single wind farm and only fine-tuned for the other wind farms. Even though the autoencoders were initially only trained on a single wind farm, the evaluation results show improved results compared to similar techniques.

In~\cite{Zhang2018}, the authors, use inductive transfer learning to classify iced wind turbines. First, a neural network is trained on a single wind turbine \textit{B}, and then it is finetuned on a group of wind turbines \textit{A}. Afterward, the classifier is tested on a remaining group of 99 wind turbines from group \textit{A}. The repeated experiment for 100 different combinations, shows an improvement of 14\% compared to an oversampling strategy.

\cite{Tasnim2018} proposes a cluster based \ac{MSDA} approach. The approach allows to cluster similar wind data distributions based on historical weather records from multiple sites. These wind distributions allow fusing predictions from multiple source sites, utilizing a weighting scheme, for a new wind farm. The cluster-based \ac{MSDA} achieved an improvement of $20.63\%$ compared to a traditional \ac{MSDA} approach.

\cite{GS16, GS18} present the \ac{CSGE} in the context of renewable power forecasts. It comprises a hierarchical two-stage ensemble prediction system and weights the ensemble member's predictions based on three aspects, namely global, local, and time-dependent performance. \cite{Deist2018} shows that it is utilizable with an arbitrary cost function. The results in \cite{Deist2018} allow to implement the \ac{CSGE} in the context of \ac{TL} with a suitable cost function, as detailed later on.

In summary, to the best of our knowledge, there is yet little research towards an evaluation on the applicability of \ac{TL} in the field of renewable energy, that allows reutilizing data and models from other wind farms throughout the whole lifecycle from its first operation on the electrical grid.

\section{Research Proposal}
\label{resproposal}

The following sections provide details of the research proposal based on the previously mentioned research questions. By solving three import {\wpl}s, we evaluate the applicability of \ac{TL} in the field of renewable energy. Solving these {\wpl}s allow reutilizing data and models from other wind farms throughout the whole lifecycle from a wind farm's first operation on the electrical grid answering question one and two. Besides, the developed methods are capable to gradually integrate an increasing amount of data answering the research question three.  
To develop these methods and provide a solution we identified the three {\wpl}s that are relevant to tackle from an \ac{ML} perspective as well for renewable energies.

\wpu~1 in \SEC{prop_no_data} discusses problems and solutions that arise when there is no historical power data about a wind farm showing strategies to select models and data from other wind farms in the absence of no historical power data; answering the research question one and two. \wpu~2 in \SEC{prop_little_data} discusses problems that arise when there is limited historical power data about a wind farm; answering the research question one and two. However, it also shows the potential improvements from the additional historical data. \wpu~3 in \SEC{prop_trans_data} deals with problems arising from the addition of larger historical data sets. In particular, it looks at how a change in input data compared to old data affects estimates of the wind power forecast. Also, it considers what volume of historical data is required so that traditional \ac{ML} instead of \ac{TL} approaches can be applied. Respectively this section answers all three research question with a strong focus on the third question.

\subsection{\wpt~1: Providing Day-Ahead Forecasts of the Expected Power Generation without Historical Power Data}
\label{prop_no_data}

Typically, when there is no historical power data of a new build wind farm, we need to use the physical model to provide a forecast of the expected power generation. The model has the rotor area, forecasted wind speed, and forecasted air density from a \ac{NWP} as inputs to provide day-ahead estimates of the expected power generation. These estimates of the model are used to devise decisions regarding costs and risks in the electrical grid. 

\begin{center}
    \begin{figure}
        \label{fig:no_data_approach}
        \centering
        \includegraphics[width=0.7\textwidth]{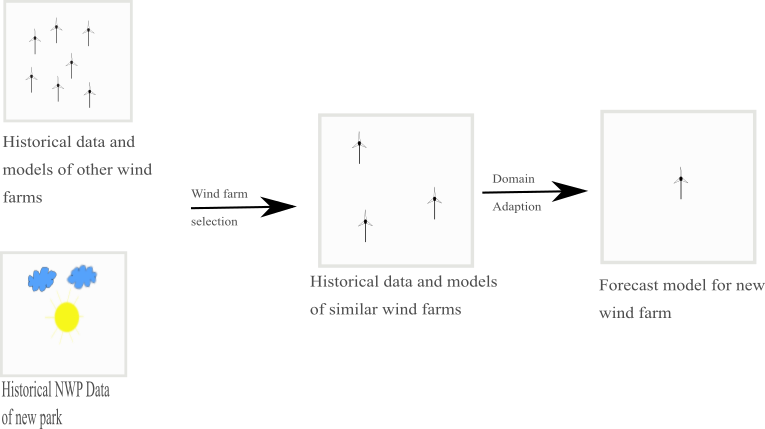}
        \caption{Scheme depicting the procedure to use domain adaption for a new wind farm. First, comparable wind farms are identified through similar weather conditions or terrain. Second, model and historical data of similar parks are adapted to be applicable for the new wind farm.}
    \end{figure}
\end{center}

However, even though there is no historical power data for a new site, typically information from the source in the form of historical $D_S$ knowledge and models $T_S$ from other wind parks are available. This \textit{knowledge} $D_S$ and $T_S$ allows providing a more sophisticated forecast in the target domain compared to the physical model. This information of data and models is adapted from the source domain to the target domain with \ac{TL} techniques. Further on we typically can extract historical \ac{NWP} $D_T$ about the location of the new wind farm and has information about the location's terrain (onshore, offshore, forest, farmland, and central mountain zone). All these pieces of information can be used to select relevant models and historical data from other wind farms as detailed later on. Respectively, in this \wpl~the following difficulties arise:

\begin{enumerate}
    \item Historical data $X_S$ and $Y_S$ from other wind farms is probably distorted towards the specifics of its original location. E.g., turbulence caused by surrounding trees influence an onshore wind farm located in the forest. These turbulences are different compared to effects occurring in offshore wind farms \cite{Schreiber2018}.

    \item Similar to the historical data, forecast models $T_S$ are probably biased towards specifics of a particular wind farm \cite{Pan2010}. Further on, the parameters of the machine learning models are biased towards the data of the original wind farm and may be hard to adapt for the new wind farm.
\end{enumerate}
To cope with these challenges, we aim to solve it through a two-step approach as shown in \FIG{fig:no_data_approach}. Step 1 automatically configures a suitable selection of historical data and forecast models that are most similar to the new wind farm. This \textit{self-configuration} can be made by analyzing influences in forecast models and derive respective rules that allow identification based on the location or historical NWP data. The location-based similarity identification is a pre-selection strategy for the second step. It minimizes the risk of distortion towards the specifics of the source domain Step 2 derives a model based on pre-selected historical data and models using \ac{TL} approaches.

One naive approach for Step 2 is to artificially create historical wind power data for the new wind farm. The pre-selected forecast model can do this with \ac{NWP} from the new wind farm's location as input. Afterward, we can train a machine learning algorithm for the new wind farm with the historical NWP data, and the artificially created power data. Another method is to create a high-dimensional representation of the NWP data for the new wind farm using an autoencoder and then use a universal forecast model trained on the historical data of similar wind farms. These two semi-supervised approaches will provide a self-optimization scheme for new wind farms.

\vspace{-4em}

\subsection{\wpt~2: Providing Day-Ahead Forecasts of the Expected Power Generation with little Historical Power Data}
\label{prop_little_data}    

Similar to the \wpl~from \SEC{prop_no_data}, we aim at developing an \ac{ML} model that allows day-ahead forecasts. In particular, we are interested in the scenario with limited $Y_T$ of a new wind farm. This historical data includes observations from, e.g., the first month after the beginning of the wind farm's first operation. Ideally, this data allows developing a model that improves the forecast quality over the physical model as well as other models developed in \SEC{prop_no_data}. Even though the problem is similar to the one in \SEC{prop_no_data}, the additional $Y_T$ has different challenges as well as potential solutions:

\begin{enumerate}
    \item One problem is when the new wind farm has a characteristic curve that is not part of the historical wind farm pool. This mismatch can happen due to a false pre-selection strategy of the similar wind farms for example due to a yet unknown wind turbine type or topology.
    \item The few historical data of the first month might be biased towards specific weather conditions only applicable to this month. Similar problems occur in case initial maintenances at the wind farm distort the first month's data. 
\end{enumerate}

What the described challenges above have in common is that the model could learn a relationship between input and output that is specific to the historical data. Consequently, the model will not generalize well for the next's month data. In the worst scenario, the data causes negative transfer, so that the new model is worse than the one from similar wind farms or the one from \SEC{prop_no_data}. 

Similar to \SEC{prop_no_data}, we first use the pre-selection strategy to configure the pool of relevant wind farms. Based on this pool we propose three potential solutions for Step 2 in \FIG{fig:no_data_approach}. The first one is to use an ensemble technique such as the \acf{CSGE} that provides a self-configuration and self-optimization scheme for Step 2 of the \wpl. The \ac{CSGE} has a so-called \textit{local} and \textit{global} weighting scheme. The global weighting scheme provides an assessment strategy to evaluate and weight the pre-selected models on the historical data of the first month. In a sense, here, this can be used to derive a weighting on how similar the models are to the new wind farm even with few historical power data. At the same time, regardless of the global weighting, the individual models provide forecasts for data not yet measured at this location to avoid negative transfer. The local weighting scheme allows deriving a weighting based on similar weather situations. This local weighting probably further mitigates the limited amount of historical power data. Further on, by adding a model to the ensemble that is trained only on the historical data of the new farm, we can avoid the first challenge, because its performance will be taken into account by the global and the local weighting scheme for the unknown characteristic curve. 

The second approach is to use a combination of finetuning and the so-called self-training approach. Here, first, a generic model based on the pool of all similar wind farms is trained. This model is then finetuned based on the historical data for the first month. The resulting model is then used to create forecasts of the power for historical NWPs of the new wind farm. The artificially created historical wind power allows to further train the initial finetuned model based on confident prediction (e.g., $90\%$) until a specified stopping criterion is met. The combination of artificially creating historical data and retraining of the model is called self-training. By training the model on the artificial data, the self-training approach improves the model for other weather conditions. Further on, by just selecting data with high confidence the self-training avoids negative transfer for unknown data.

The third approach is to use an \ac{MTL} strategy based on hard-parameter sharing. We first learn a model based on the pre-selected pool of similar wind farms. In the multi-task approach, all wind farms are trained at the same time. This simultaneous training allows the algorithm to derive a universal representation of the NWP data. Once, the model is trained we add the task for the new wind farm and finetune the model. Due to the generic nature of the \ac{NWP} representation, only a small amount of historical data is needed to adapt the weights for the new task. Because of a large amount of data for the initial training, typically the generic representation allows deriving sophisticated models even with few historical data and provides a suitable self-optimization framework. Further, by the adaption strategy based on the one month of training data, we assure that the algorithm learns the specifics of the new wind farm compared to the universal representation. The combination of historical data specific to the wind farm and the generic representation avoids the negative transfer by only learning the relevant data valid to the farm.

\subsection{\wpt~3: Providing Day-Ahead Forecasts of the Expected Power Generation with Increasing Historical Data and Lifetime}
\label{prop_trans_data}
While in \SEC{prop_little_data} there was little historical data, with increasing lifetime of a wind farm more and more data $Y_T$ is collected. Based on the model created in \SEC{prop_little_data} we now need to integrate this data into the algorithms gradually. The continuous integration of new historical data $Y_T$ increases the forecast performance. However, a change in the data compared to past data affects estimates of the wind power forecast. This change in data can be related to a newly integrated night shut-off, maintenance, or the implementation of a different weather model. Respectively, this \wpl~has the following difficulties:

\begin{enumerate}
    \item How to detect and adapt the model for unknown situations such as maintenance?
    \item How to minimize the training effort concerning computational costs for conceptual drifts in the data, e.g., from a new weather model?
\end{enumerate}

Importantly to note, the selection from the pool of similar wind farms is here potentially different to the other {\wpl}s. This difference is because previously we had little information about the farm and were only considering data from similar locations and characteristics. However, now we are interested in similar situations. In the case of a night shut-off, the data is identical for all wind farms that implement it. In all implementations, the generated power will be zero within a certain time span. 

Recognizing such changes is typically solved with anomaly or novelty detection \cite{Gru16}. Afterward, we can use a clustering algorithm such as k-nearest neighbors to find similar training data from other wind farms. After finding the related pieces of training data, we are again able to use finetuning and self-training to adapt the original model to the novel situations as described in \SEC{prop_little_data}. By using finetuning, we can minimize computational costs compared to traditional \ac{ML} approaches.

An iterative process of the \ac{CSGE} gives a more automatic approach. Consider we have a \ac{CSGE}, where we combine models of all wind farms. With an increasing amount of data for the new wind farm, the individual model's performance will increase and respectively its weighting compared to other wind farms. However, now consider the installation of night shut-off. Once, we retrain the \ac{CSGE} with data containing the night shut-off, the global performance of the novel wind farm model will decrease while the models with knowledge about night shut-off will increase. This retraining of the CSEGE allows adapting to novelties automatically. Additionally, one also needs to finetune the new wind farm model based on the new data, to assure that the model learns to predict night shut-offs or similar scenarios over time.

An \ac{MTL} approach tackles the scenario with a new NWP model. Consider a neural network, where multiple NWP models are the input for the \ac{MTL} model. Further, the model is trained so that the power generation forecast is independent of a specific weather models input. This approach will force the algorithm to learn a generic representation of the NWP models. Respectively, allowing to apply domain adaption techniques for a new \ac{NWP} model.

\begin{center}
    \begin{figure}
        \centering
        \label{multi_cross_learning}
        \includegraphics[width=0.75\textwidth]{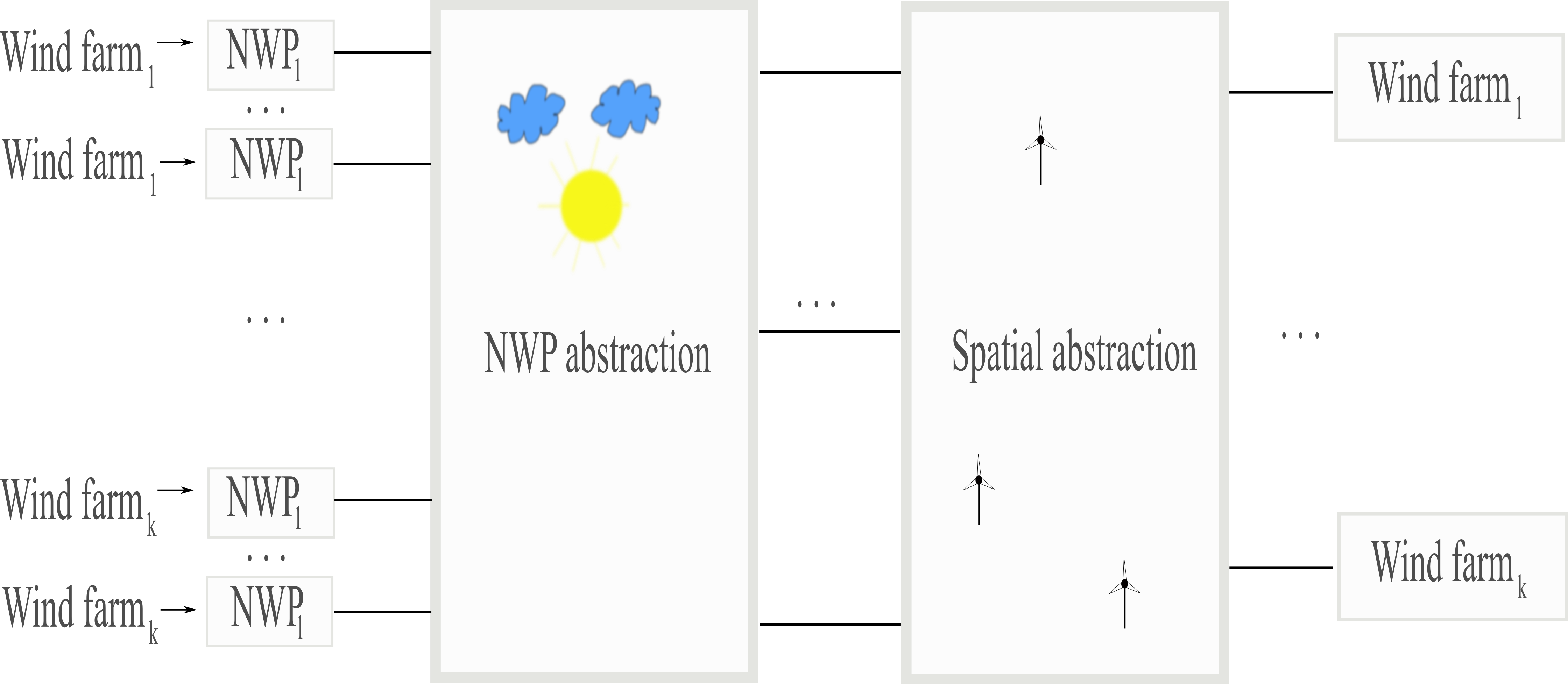}     
        \caption{Multi-task-learning, which serves to transfer knowledge about numerical weather models as well as to model the dependencies between individual wind parks and their prediction models. Multiple \ac{NWP} models are the input for the \ac{MTL} model. Further, the model is trained so that the power generation forecast is independent of a specific weather model's input.} 
    \end{figure}
\end{center}

This idea can be further extended with a spatial abstraction layer, see \FIG{multi_cross_learning}, to what we call \textit{multi-cross-learning}. Initially, during training, the model learns with data from $l$ weather models and $k$ wind farms at the same time. The simultaneous training causes the model to determine a generic representation of the NWP features on the one hand and of the spatial relation of the wind farms on the other side. The resulting model allows tackling various problems at once. In case we are faced with novelty for wind farm $i$, domain adaption of the spatial abstraction will enable to use models from other wind farms. Further, this allows integrating new NWP models easily. The generic abstraction of the NWP allows limiting the training to a transformation between the new NWP model and the NWP abstraction. Finally, this approach can be considered as an extension of the self-optimization scheme from the traditional \ac{MTL} approach by providing additional layers of abstraction.

\vspace{-4em}

\section{Progress Description and Summary}
\label{sec:prog_desc}

In this section, we summarize our preliminary work and outline the research proposal. Preliminary work in~\cite{Schreiber2018} provides a method that analyzes the influence of so-called probabilistic forecasts method. The results of this method will be used later on to design the pre-selection strategy as described in \SEC{prop_no_data} and~\ref{prop_little_data}. Therefore in~\cite{Schreiber2018}, a hybrid feature selection strategy is implemented, and influences are analyzed with the so-called sensitivity analysis. The results show that, as expected, wind speed at 100m altitude is the essential feature in all evaluated terrains and \ac{ML} models. Similar, in most cases wind speed at 10m height has the second most significant influence. If selected, air pressure and the variability of wind speed at 100m altitude have about the third highest impact for all evaluated scenarios.

In~\cite{Deist2018}, we proposed the \ac{CSGE} for general machine learning tasks and interwoven systems. This paper shows that the \ac{CSGE} can be optimized according to arbitrary loss functions making it accessible for a broader range of problems and providing a self-improving scheme based on previously seen data. This self-improving scheme can be applied to the self-configuration and self-optimization scenarios as proposed in \SEC{prop_little_data} and~\ref{prop_trans_data}. Moreover, we showed the applicability and easy interpretability of the approach for synthetic datasets as well as real-world data sets. For the real-world datasets, we showed that our \ac{CSGE} approach reaches state-of-the-art performance compared to other ensembles methods for both classification and regression tasks.

Conclusively, in this \atl, we proposed various techniques that allow utilization of transfer learning techniques in the field of renewable power forecasts. The proposal is the first step toward a broader and automated use of transfer learning in other domains besides \ac{NLP} and vision. We are currently in the process of investigating the multi-cross task learning approach and the superior forecast quality of transfer learning over the physical model. Initial findings show the applicability and motivate further research in this area.

\ifthenelse{\boolean{article}}{}
{
\section{Time Plan}
\label{timeplan}

The time plan shows the expected person months and the dependencies between the \wpl. 
% \begin{center}
\begin{figure}

\noindent\resizebox{.9\textwidth}{!}{
  \begin{ganttchart}[vgrid, hgrid]{1}{30}
    \gantttitle{Time in month}{30} \\
    \gantttitlelist{1,...,30}{1} \\
    \ganttbar{WP 1}{1}{6} \\ %\ref{sec:class}
    \ganttbar{WP 2}{7}{14} \\ %\ref{sec:activ}
    \ganttbar{WP 3}{15}{23} \\ %\ref{sec:spec}
    \ganttbar{Evaluation}{1}{26} \\ %\ref{sec:case}
    \ganttbar{Writing}{4}{30}
  \end{ganttchart}
}
\caption{Time Plan of the Three {\wpt}s Including the Expected Time in Month for Evaluation and Writing of the Thesis.}  
\end{figure}
}
% \end{center}
\vspace{-1em}
\bibliographystyle{ieeetr}
\bibliography{literature}

% that's all folks
\end{document}